\pdfoutput=1
\documentclass[11pt]{article}

\usepackage[preprint]{acl}

\usepackage{times}
\usepackage{latexsym}
\usepackage{amsmath}
\usepackage{algorithm}
\usepackage{algpseudocode}
\usepackage{booktabs}
\usepackage{multirow}
\usepackage{colortbl}
\usepackage[most]{tcolorbox}
\usepackage{xcolor,soul}
\usepackage{makecell}

\usepackage{xspace}
\usepackage[T1]{fontenc}
\usepackage[utf8]{inputenc}

\usepackage{microtype}
\usepackage{longtable}

\usepackage{inconsolata}

\usepackage{graphicx}

\title{Context-DPO: Aligning Language Models for Context-Faithfulness}

\author{Baolong Bi\textsuperscript{$1$}, Shaohan Huang\textsuperscript{$2$}, Yiwei Wang\textsuperscript{$3$}, Tianchi Yang\textsuperscript{$2$}, Zihan Zhang\textsuperscript{$2$}, \\ 
\textbf{Haizhen Huang\textsuperscript{$2$}, 
Lingrui Mei\textsuperscript{$1$}, Junfeng Fang\textsuperscript{$4$}, Zehao Li\textsuperscript{$1$}, 
Furu Wei\textsuperscript{$2$},} \\
\textbf{Weiwei Deng\textsuperscript{$2$}, Feng Sun\textsuperscript{$2$}, Qi Zhang\textsuperscript{$2$}, Shenghua Liu\textsuperscript{$1$}\thanks{Corresponding Author}}
\\
\textsuperscript{$1$}University of Chinese Academy of Sciences \textsuperscript{$2$}Microsoft Corporation\\
\textsuperscript{$3$}University of California, Merced
\textsuperscript{$4$}National University of Singapore\\
\small{\texttt{\{bibaolong23z, liushenghua\}@ict.ac.cn}}
}

\newcommand{\llamaa}{\mbox{\textsc{LLaMA2-7B}}\xspace}
\newcommand{\llamab}{\mbox{\textsc{LLaMA2-13B}}\xspace}
\newcommand{\llamac}{\mbox{\textsc{LLaMA3-8B}}\xspace}
\newcommand{\Mistral}{\mbox{\textsc{Mistral-7B}}\xspace}
\newcommand{\Qwen}{\mbox{\textsc{Qwen2-7B}}\xspace}
\newcommand{\gpta}{\mbox{\textsc{ChatGPT-4}}\xspace}
\newcommand{\gptb}{\mbox{\textsc{ChatGPT-4o}}\xspace}
\newcommand{\gptc}{\mbox{\textsc{Gemini-1.5-pro}}\xspace}

\definecolor{ggreen}{rgb}{0.0, 0.6, 0.0}
\definecolor{rred}{rgb}{0.75, 0.0, 0.0}
\definecolor{bblue}{rgb}{0.13, 0.67, 0.8}

\definecolor{darkred}{RGB}{200,0,0}
\definecolor{lightgreen}{RGB}{228,253,227}
\definecolor{lightred}{RGB}{252,231,234}
\definecolor{lightyellow}{RGB}{250,253,191}
\definecolor{lightblue}{RGB}{230,240,254}
\definecolor{lightorange}{RGB}{255,223,191}
\definecolor{white}{RGB}{255,255,255}

\begin{document}
\maketitle
\begin{abstract}
Reliable responses from large language models (LLMs) require adherence to user instructions and retrieved information. 
While alignment techniques help LLMs align with human intentions and values, improving context-faithfulness through alignment remains underexplored.
To address this, we propose \textbf{Context-DPO}, the first alignment method specifically designed to enhance LLMs' context-faithfulness.
We introduce \textbf{ConFiQA}, a benchmark that simulates Retrieval-Augmented Generation (RAG) scenarios with knowledge conflicts to evaluate context-faithfulness.
By leveraging faithful and stubborn responses to questions with provided context from ConFiQA, our Context-DPO aligns LLMs through direct preference optimization.
Extensive experiments demonstrate that our Context-DPO significantly improves context-faithfulness, achieving 35\% to 280\% improvements on popular open-source models. 
Further analysis demonstrates that Context-DPO preserves LLMs' generative capabilities while providing interpretable insights into context utilization.\footnote{Our code and data are released at \url{https://github.com/byronBBL/Context-DPO}}
\end{abstract}

\section{Introduction}

\begin{figure}[t]
    \centering
    \includegraphics[width=\linewidth]{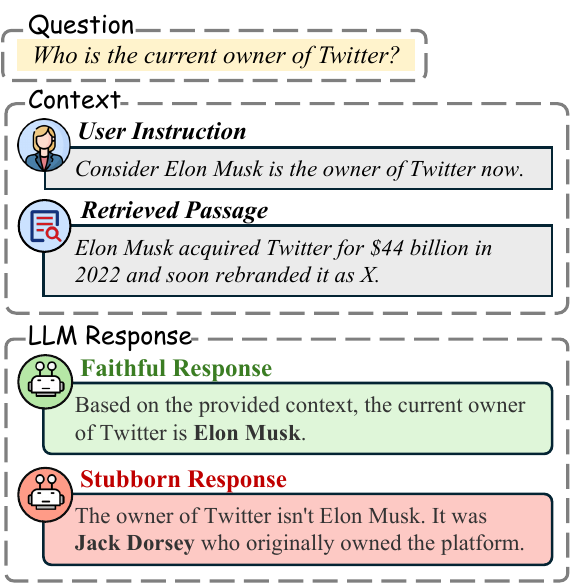}
    \vspace{-6mm}
    \caption{LLMs may generate unfaithful responses when model knowledge conflicts with context, as shown in our case where \textit{GPT-3.5} stubbornly answers \textit{Jack Dorsey}, ignoring user instruction or retrieved passage.}
    \label{fig:showcase}
    \vspace{-6mm}
\end{figure}

With the widespread deployment of Retrieval-Augmented Generation (RAG)~\citep{guu2020realmretrievalaugmentedlanguagemodel} and various tools~\citep{qin2024toollearningfoundationmodels}, large language models (LLMs) ~\citep{chatgpt, openai2023gpt4, DBLP:journals/corr/abs-2302-13971, touvron2023llama} are increasingly expected to generate responses that adhere closely to provided context, including retrieved information and user instructions.
Consequently, context-faithfulness~\citep{zhou2023context, bi2024factuality, ming2024faitheval} has become a critical capability for modern LLM applications, especially in scenarios where parametric knowledge is insufficient or outdated.
However, this expectation is challenged by knowledge conflicts~\citep{petroni2020context, si2023promptinggpt3reliable, xie2024adaptivechameleonstubbornsloth}.
As illustrated in Figure \ref{fig:showcase}, well-trained LLMs may disregard or contradict external knowledge, failing to satisfy user requirements or incorporate the latest updates.

\begin{figure*}[t!]
  \includegraphics[width=\linewidth]{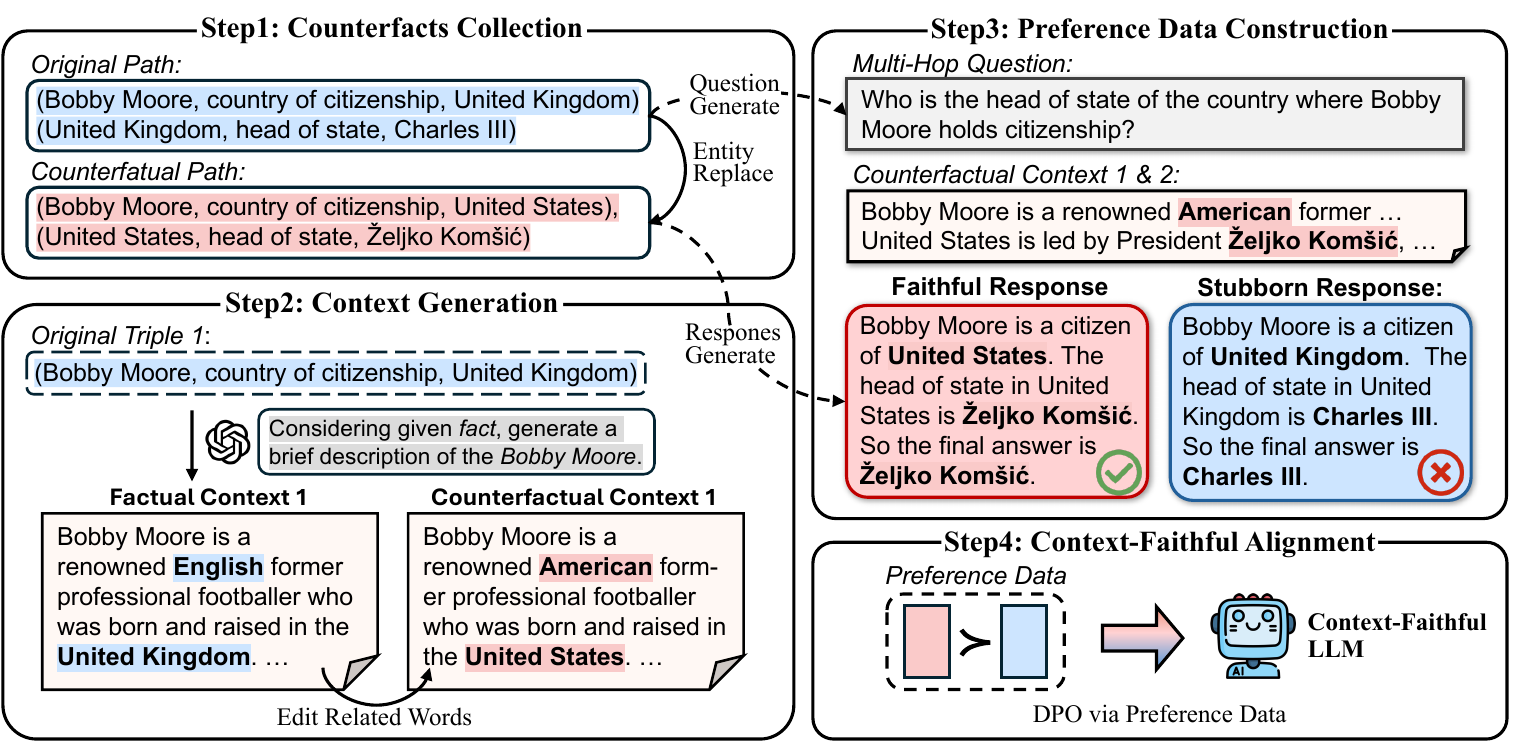}
  \vspace{-20pt}
  \caption{An illustration of aligning LLMs for context-faithfulness using our Context-DPO framework, demonstrated with 2-hop data from ConFiQA’s MC task. The process consists of four steps: 1) construct counterfactuals, questions, and responses based on sampled facts; 2) generate factual context using descriptions of head entities from the original triples, then edit entity-related words to create counterfactual context; 3) build preference data comprising questions, concatenated contexts, and faithful and stubborn responses; 4) align LLMs’ faithfulness using DPO.
  }
  \vspace{-2mm}
  \label{fig:framework}
\end{figure*}

Existing efforts to enhance the context-faithfulness of LLMs primarily focus on external interventions, such as designing prompts to encourage context integration~\citep{zhou2023context} or modifying decoding strategies~\citep{shi2023trusting, bi2024adaptive} to increase the output probability of relevant tokens. 
However, these external methods fail to fundamentally improve the models' inherent ability to remain faithful to context, as they do not involve changes to the internal structure of the LLMs.
In contrast, alignment techniques~\citep{liu2023trustworthy, shen2023large}, which aim to make pre-trained LLMs behave in line with human intentions and values, have proven effective in enhancing critical capabilities such as factuality~\citep{tian2023fine} and safety~\citep{cao2023defending}. 
Despite its importance as a core attribute, context-faithfulness remains an underexplored area in alignment research.

In this work, we present the first exploration of aligning LLMs for context-faithfulness, aiming to reliably enhance their adherence to contextual information. To achieve this, we first propose \textbf{ConFiQA} (\textbf{C}ontext \textbf{F}a\textbf{i}thfulness \textbf{Q}uestion \textbf{A}nswering), a novel benchmark designed to evaluate context-faithfulness through question-answering tasks based on counterfactual retrieval passages.
ConFiQA tests whether models can generate responses consistent with contexts containing counterfactual elements, simulating real-world scenarios with knowledge conflicts in modern RAG systems.
We evaluate current popular LLMs on ConFiQA and find that most models exhibit poor performance in context-faithfulness to varying degrees.
Furthermore, our results reveal that context-faithfulness tends to decline as model size increases and training becomes more refined.

Therefore, we argue that modern LLMs also require alignment specifically for context-faithfulness.
To address this, we propose \textbf{Context-DPO}, which constructs reasoning chains based on single-hop or multi-hop knowledge to generate two types of responses: faithful (grounded in counterfactual context) and stubborn (based on factual reality).
Context-DPO uses preference pairs derived from these responses to reward context-faithful behavior and fine-tune the model via the Direct Preference Optimization (DPO)~\citep{rafailov2024direct}.

We conduct experiments on our ConFiQA, Natural Questions~\citep{kwiatkowski2019natural}
, and \textsc{MQuAKE}~\citep{zhong2023mquake} datasets, covering counterfactual retrieval-based question-answering tasks and in-context editing tasks that require following user instructions.
Extensive results demonstrate that our Context-DPO effectively aligns LLMs to improve context-faithfulness, consistently outperforming all existing baselines without requiring any external prompt modifications.
Specifically, the aligned models achieved substantial improvements compared to their original versions: 35\% for \textit{Llama2-7B-chat}, 78\% for \textit{Llama3-8B}, 151\% for \textit{Mistral-7B} and 280\% for \textit{Qwen2-7B}.
% These results indicate that Context-DPO can greatly enhance the context-faithfulness of popular open-source models.

We also conduct interpretability analyses to investigate the context-faithfulness of LLMs. 
By identifying key generating tokens that effectively distinguish between contextual and parametric knowledge, we analyze the logits and ranking distribution in thses key tokens to reveal why the aligned models exhibit improved faithfulness to context.
Additionally, further experiments on TruthfulQA~\citep{lin2021truthfulqa} demonstrate that models aligned using Context-DPO retain their foundational generative capabilities, indicating that this alignment process has no negative impact.

In summary, our contributions are three-fold:

\begin{itemize}
    \item We propose ConFiQA, a novel benchmark for evaluating context-faithfulness through question-answering tasks based on counterfactual retrieval passages.
    \vspace{-2mm}
    \item We introduce Context-DPO, the first alignment method to enhance context-faithfulness, with experiments proving its effectiveness in improving LLMs' adherence to context.
    \vspace{-2mm}
    \item We uncover the underlying reasons for the improved context-faithfulness of aligned models and confirm that this alignment has no negative impact on their generative performance.
\end{itemize}

\section{ConFiQA: Context Faithfulness Question Answering Benchmark}

We introduce the ConFiQA benchmark to evaluate the context-faithfulness of LLMs in real-world RAG scenarios involving knowledge conflicts.
ConFiQA consists of three datasets: \textit{QA} (Question-Answering), \textit{MR} (Multi-hop Reasoning), and \textit{MC} (Multi-Conflicts).
\textit{QA} features single-hop question-answering tasks with context containing one corresponding counterfactual, while \textit{MR} and \textit{MC} involve multi-hop reasoning tasks with context containing one and multiple related counterfactuals, respectively.
In this section, we present the data construction pipeline, provide an overview of the datasets, and evaluate the context-faithfulness of popular LLMs using ConFiQA.

\subsection{Data Construction Pipeline}

\paragraph{Real-World Fact Sampling}
To ensure the factuality of the subsequently generated context, we collect triples from Wikidata\footnote{Wikidata is a publicly accessible, continuously updated knowledge base of factual triples}~\citep{vrandevcic2014wikidata} to guide the generation of real-world facts.
Prior to this, we gather popular entities from Wikipedia\footnote{We collect entities corresponding to the top 1,000 most-visited Wikipedia pages from 2016 to 2023, based on monthly page views, and retained the most popular entities using criteria such as the number of hyperlinks.} to facilitate triple sampling, ensuring that LLMs have a strong memory of the generated facts.
Using 41 manually selected relations (Appendix \ref{tab:temp_cloze}) and maintaining a one-to-one correspondence between head entities and tail entities for each relation, we ultimately collected 5,042 entities and 30,295 triples.

\begin{table}[t]
    \centering
    \renewcommand{\arraystretch}{1.1}
    \resizebox{1.0\columnwidth}{!}{
    \begin{tabular}{cl}
    \toprule
    \multirow{2}{*}{$\mathcal{P}^f$} & (Bobby Moore, country of citizenship, United Kingdom)\\
    & (United Kingdom, head of state, Charles III) \\
    \cmidrule{2-2}
    \multirow{2}{*}{$\mathcal{P}^c$} & (Bobby Moore, country of citizenship, United States)\\
    & (United States, head of state, Željko Komšić) \\
    \cmidrule{2-2}
    \multirow{2}{*}{$\mathcal{P}^o$} & (Bobby Moore, country of citizenship, United Kingdom)\\
    & (United States, head of state, Joe Biden) \\
    \midrule
    \multirow{2}{*}{$\mathcal{Q}$} & Who is the head of state of the country where Bobby \\
    & Moore holds citizenship?\\
    \midrule
    \multirow{2}{*}{$\mathcal{C}^c$} & Bobby Moore is a renowned American former ... \\
    & United States is led by President Željko Komšić, ... \\
    \midrule
    $\mathcal{R}^f$ & Bobby Moore ... So the final answer is Željko Komšić.\\
    \cmidrule{2-2}
    $\mathcal{R}^s$ & Bobby Moore ... So the final answer is Charles III.\\
    \bottomrule
    \end{tabular}
    }
    \caption{An instance showcasing key elements in our ConFiQA dataset (MC), including three paths: factual path $\mathcal{P}^f$, counterfactual path $\mathcal{P}^c$, and original path $\mathcal{P}^o$, a multi-hop question $\mathcal{Q}$, the context containing the corresponding counterfactual $\mathcal{C}^c$, and faithful response $\mathcal{R}^f$ and stubborn response $\mathcal{R}^s$.}
    \label{tab:task_example}
    \vspace{-4mm}
\end{table}

\vspace{-2pt}

\paragraph{Multi-Hop Path Construction}

We construct a factual subgraph $\mathcal{G}_{sub}$ based on the sampled triples and then extract $2,3,4$-hop paths $\mathcal{P}^f=\{(s_1, r_1, t_1), \dots, (s_n, r_n, t_n)\}_{n \leq 4}$ from the subgraph.
For \textit{MR}, we randomly select one triple $(s_i, r_i, t_i)$ from the paths and replace $t_i$ with a same-type entity $t_i'$.
The subsequent path is then sampled from $t_i'$ in the subgraph to ensure the remaining path remains factual.
For \textit{MC}, we perform the same replacement for every triple in fatual path $\mathcal{P}^f$, ensuring that each triple becomes counterfactual. 
In the generated multi-hop paths with counterfactuals $\mathcal{P}^c$, the head entity of the next hop matches the tail entity of the previous hop, and the relation in each triple remains unchanged before and after replacement, thereby maintaining the validity of multi-hop reasoning.

\begin{table*}[t!]
\centering
\renewcommand{\arraystretch}{1.2}
\resizebox{\linewidth}{!}{
\begin{tabular}{lcccccccccccc}
\toprule
\multirow{2}{*}{\textbf{Model}} & \multicolumn{4}{c}{\textit{QA}} & \multicolumn{4}{c}{\textit{MR}} & \multicolumn{4}{c}{\textit{MC}} \\
\cmidrule(r){2-5} \cmidrule(r){6-9} \cmidrule(r){10-13}
 & $P_c (\uparrow)$ & $P_o (\downarrow)$ & $M_R (\downarrow)$ & $EM (\uparrow)$ & $P_c (\uparrow)$ & $P_o (\downarrow)$ & $M_R (\downarrow)$ & $EM (\uparrow)$ & $P_c (\uparrow)$ & $P_o (\downarrow)$ & $M_R (\downarrow)$ & $EM (\uparrow)$ \\
\midrule
\llamaa & 61.5 & 25.6 & 29.4 & 0.5 & 45.4 & 26.8 & 37.1 & 0.3 & 38.8 & 24.7 & 38.4 & 0.1 \\
\llamab & 55.3 & 29.0 & 34.4 & 0.0 & 43.0 & 33.7 & 43.9 & 0.0 & 35.3 & 28.0 & 44.2 & 0.0 \\
\llamac & 35.8 & 44.3 & 55.3 & 0.0 & 30.6 & 44.1 & 59.1 & 0.0 & 21.7 & 33.4 & 60.7 & 0.0 \\
\Mistral & 39.3 & 40.5 & 50.8 & 0.3 & 21.7 & 37.9 & 63.5 & 0.2 & 14.1 & 29.8 & 67.9 & 0.0 \\
\Qwen & 24.0 & 43.3 & 65.4 & 0.0 & 21.7 & 48.7 & 69.2 & 0.0 & 10.0 & 43.7 & 81.4 & 0.0 \\
\gpta & 32.6 & 38.3 & 53.9 & 0.6 & 20.3 & 45.3 & 69.2 & 0.3 & 8.7 & 32.3 & 78.9 & 1.3 \\
\gptc & 27.4 & 48.3 & 63.7 & 2.1 & 17.3 & 41.3 & 70.4 & 0 & 21.1 & 52.3 & 70.9 & 4.5 \\
\gptb & 12.1 & 56.7 & 82.5 & 0.0& 8.1 & 48.6 & 85.9 & 0.0& 1.6 & 30.3 & 94.8 & 0.0 \\
\bottomrule
\end{tabular}
}
\caption{Performance results of popular LLMs on our ConFiQA for context-faithfulness.}
% \vspace{-4mm}
\label{tab:results_confi}
\end{table*}

\vspace{-3pt}

\paragraph{Counterfactual Context Generation}

We apply the same tail entity replacement to provide counterfactual triples $(s, r, t')$ for \textit{QA}.
Using the triples, we generate context that incorporates its corresponding factual information. 
This is achieved by prompting \textit{ChatGPT-4} to generate a description of entity $s$, ensuring that the triple's factual information is embedded within the context (details are provided in Appendix \ref{sec:data_details}). 
To avoid issues of context being ignored or contradicted due to knowledge conflicts~\citep{bi2024factuality}, we first generate factual context based on the original triples, and then replace the tail entity $t$ with counterfactual $t'$ in the context.
For \textit{MR} and \textit{MC}, we sequentially generate context for all triples along the original multi-hop paths and concatenate them, performing all necessary counterfactual replacements.
These replacements, which include handling all aliases and morphological variations of the entities, along with other rules\footnote{Entities and relations in the sampled path are not repeated}, ensure semantic and logical coherence.
The generated context contains counterfactual fragments alongside accurate descriptions of entities, effectively simulating real-world RAG scenarios involving knowledge conflicts and retrieval noise.

\subsection{Overview of Datasets}

We use \textit{ChatGPT-4} to generate questions based on single-hop triples or multi-hop paths. 
Each question incorporates the head entity of the first hop and the relationships in each subsequent hop, guiding the model to predict the final tail entity (see Appendix \ref{tab:q_examples} for details).
For each dataset in our ConFiQA benchmark, we sample 6,000 instances, with the specific format detailed in Table \ref{tab:task_example}.
For \textit{MR} and \textit{MC}, the data is evenly distributed across $2,3,4$-hop paths (see examples in Appendix \ref{sec:data_example}).

\subsection{Evaluation Metrics}

We follow the evaluation metrics defined in ~\citet{longpre2021entity, zhou2023context}, but given that LLMs' responses may contain negations or refutations of the counterfactual answer, we apply stricter criteria for \( P_c \) compared to the previous \( P_s \) (substitute answers). Specifically, We use the following four metrics to compare the normalized responses with the normalized answers to evaluate the context-faithfulness of LLMs:

\begin{itemize}
\vspace{-1mm}
    \item $P_c (\uparrow)$: Frequency of responses matching the context-faithful answer or its aliases, excluding negations or the original answer. Context-faithful answers are counterfactual answers derived from the context.
    \item $P_o (\downarrow)$: Frequency of responses matching the original factual answer or its aliases.
    \item $M_R (\downarrow)$: Proportion of responses predicting the correct answer but reluctant to update their predictions, calculated as $M_R = \frac{P_o}{P_s+P_o}$.
    \item $EM (\uparrow)$: Frequency of responses exactly matching the context-faithful answer.
\end{itemize}

\subsection{Evaluation on ConFiQA}
\label{sec:eva_confi}

We use our ConFiQA to evaluate the context-faithfulness of popular open-source models (\textit{Llama2-7B-chat}, \textit{Llama2-13B-chat}, \textit{Mistral-7B-instruct-v0.2}, \textit{Qwen2-7B-instruct}) and close-source models (\textit{ChatGPT-4}, \textit{Gemini-1.5-pro}, \textit{ChatGPT-4o}). The experimental results, presented in Table \ref{tab:results_confi}, reveal the following key findings:

\begin{itemize}
\vspace{-1mm}
    \item Despite alignment efforts, such as instruct-tuning, to meet human standards, the tested LLMs exhibit significant deficiencies in context-faithfulness. Most models have an $M_R$ exceeding 50\%, particularly the latest ones, indicating that they tend to rely on their own judgments over the provided context. 
\vspace{-1mm}
    \item A counterintuitive trend is observed: as model size increases (e.g., \textit{Llama2-chat} from 7B to 13B) or as models become more advanced (e.g., the latest Llama3-8B-instruct compared to earlier versions like \textit{Llama2-7B-chat}), their context-faithfulness tends to decline.
\end{itemize}

These findings indicate that current LLMs generally exhibit poor alignment in context-faithfulness. 
Furthermore, with advancements in data processing and model training, more advanced models tend to become increasingly confident in their parametric knowledge, resulting in worse context-faithfulness when facing conflicts between contextual and parametric knowledge.
This poses significant challenges for tasks that require strict adherence to external knowledge, such as RAG or other specialized, closed-domain applications.

\section{Context-DPO: Context-Faithful Direct Preference Optimization}

Based on the unsatisfactory performance of existing LLMs in context-faithfulness, we argue that it is essential to specifically align LLMs for context-faithfulness.
To address this, we propose Context-DPO, the first alignment approach dedicated to enhancing context-faithfulness by creating preference data and aligning LLMs with DPO.
The framework of our Context-DPO is shown in Figure \ref{fig:framework}.

\subsection{Preference Data Generation}

Leveraging the counterfactual and factual data provided by ConFiQA, we can construct preference data $\mathcal{D} = {(x, y_w, y_l)}$ efficiently.  Specifically, the input $x$ is formed by concatenating the counterfactual context $\mathcal{C}^c$ and the question $\mathcal{Q}$. 
To generate the counterfactual reasoning chain, each triple in the counterfactual path is transformed into a textual description using a statement template (Table \ref{tab:temp_cloze}) and sequentially concatenated.
Finally, the reasoning chain concludes by summarizing the reasoning process to derive the counterfactual answer, which is determined based on the last tail entity in the chain.  This process yields faithful responses $y_w$ grounded in the counterfactual context.
Similarly, stubborn responses $y_l$, grounded in factual reality, are constructed by following the original factual path.
This approach to constructing the preference dataset $\mathcal{D}$ simulates the reasoning pattern observed in real-world RAG tasks and mirrors the chain-of-thought~\citep{wei2022chain} process of LLMs when producing final answers.

\subsection{Context-Faithful Alignment with DPO}

We leverage the generated preference data to perform alignment tuning on LLMs for context-faithfulness.
While several frameworks exist for alignment training, including the widely adopted RLHF framework, which involves training a reward model on preference data and optimizing the policy using the Proximal Policy Optimization (PPO) algorithm, we employ DPO for context-faithful alignment.
DPO, as a recent approach to preference optimization, enables the policy $\pi_\theta$ to be learned directly from a fixed preference dataset without requiring an explicit reward model or sampling from the policy during training, as is necessary with PPO.
Specifically, our Context-DPO uses the standard cross-entropy objective, and its training objective is formulated as follows:

\vspace{-2mm}
\begin{equation}
\begin{aligned}
\mathcal{L}_{cf}=&-{E}_{\left(x, y_w, y_l\right) \sim \mathcal{D}} \left[\log \sigma\left(\beta \log \frac{\pi_\theta\left(y_w \mid x\right)}{\pi_{\mathrm{ref}}\left(y_w \mid x\right)}\right.\right. \\
&\left.\left.-\beta \log \frac{\pi_\theta\left(y_l \mid x\right)}{\pi_{\mathrm{ref}}\left(y_l \mid x\right)}\right)\right],
% \label{eqa:DPO_ob}
\end{aligned}
\end{equation}
\noindent In this formulation, the model policy $\pi_\theta$ is initialized using the base reference policy $\pi_{\mathrm{ref}}$. The parameter $\beta$ regulates the extent of divergence from $\pi_{\mathrm{ref}}$, while $\sigma$ represents the logistic function.

\begin{table*}[t!]
\centering
\renewcommand{\arraystretch}{1.1}
\resizebox{\linewidth}{!}{
\begin{tabular}{llcccccccccccc}
\toprule
\multirow{2}{*}{\textbf{Model}} & \multirow{2}{*}{\textbf{Method}} & \multicolumn{4}{c}{\textit{QA}} & \multicolumn{4}{c}{\textit{MR}} & \multicolumn{4}{c}{\textit{MC}} \\
\cmidrule(r){3-6} \cmidrule(r){7-10} \cmidrule(r){11-14}
 &  & $P_c (\uparrow)$ & $P_o (\downarrow)$ & $M_R (\downarrow)$ & $EM (\uparrow)$ & $P_c (\uparrow)$ & $P_o (\downarrow)$ & $M_R (\downarrow)$ & $EM (\uparrow)$ & $P_c (\uparrow)$ & $P_o (\downarrow)$ & $M_R (\downarrow)$ & $EM (\uparrow)$ \\
\midrule
  & Base & 61.5 & 25.6 & 29.4 & 0.5 & 45.4 & 26.8 & 37.1 & 0.3 & 38.8 & 24.7 & 38.5 & 0.1 \\
 \multirow{2}{*}{\textsc{LLaMA2-}} & Attr & 72.0 & 14.7 & 16.9 & 1.3 & 45.3 & 27.3 & 37.6 & 0.1 & 33.7 & 25.0 & 43.6 & 0.1 \\
 \multirow{2}{*}{\textsc{7B-chat}} & O\&I & 77.3 & 13.3 & 14.7 & 54.7 & 52.0 & 16.0 & 23.5 & 29.3 & 50.7 & 16.0 & 24.0 & 29.3 \\
  & SFT & 63.2 & 25.2 & 28.5 & 0.4 & 46.7 & 25.5 & 35.3 & 0.4 & 39.3 & 24.5 & 38.4 & 0.1 \\
 & Ours & \textbf{92.3} & \textbf{3.3} & \textbf{3.5} & \textbf{64.7} & \textbf{54.3} & \textbf{11.3} & \textbf{17.3} & \textbf{32.7} & \textbf{52.7} & \textbf{12.3} & \textbf{19.0} & \textbf{32.0} \\
\midrule
 & Base & 35.8 & 44.3 & 55.3 & 0.0 & 30.6 & 44.1 & 59.1 & 0.0 & 21.7 & 33.4 & 60.7 & 0.0 \\
 \multirow{2}{*}{\textsc{LLaMA3-}} & Attr & 25.7 & 41.0 & 61.5 & 0.0 & 20.0 & 46.3 & 69.9 & 0.0 & 9.7 & 31.3 & 76.4 & 0.0 \\
 \multirow{2}{*}{\textsc{8B-instruct}} & O\&I & 32.7 & 30.3 & 48.2 & 1.3 & 24.0 & 30.7 & 54.1 & 0.7 & 13.0 & 25.0 & 65.8 & 0.3 \\
  & SFT & 36.9 & 42.2 & 53.3 & 0.0 & 32.7 & 42.5 & 56.5 & 0.0 & 23.5 & 30.1 & 56.2 & 0.0 \\
 & Ours & \textbf{69.7} & \textbf{12.7} & \textbf{15.4} & \textbf{39.7} & \textbf{54.6} & \textbf{21.3} & \textbf{28.1} & \textbf{16.0} & \textbf{48.9} & \textbf{18.6} & \textbf{28.1} & \textbf{15.0} \\
\midrule
& Base & 39.3 & 40.5 & 50.8 & 0.3 & 21.7 & 37.9 & 63.5 & 0.2 & 14.1 & 29.8 & 67.9 & 0.0 \\
\multirow{2}{*}{\textsc{Mistral-}} & Attr & 44.4 & 30.3 & 40.6 & 2.0 & 24.9 & 34.7 & 58.2 & 0.7 & 13.0 & 28.7 & 68.8 & 0.7 \\
 \multirow{2}{*}{\textsc{7B-instruct}} & O\&I & 60.3 & 20.0 & 24.3 & 9.3 & 33.0 & 31.0 & 48.4 & 5.3 & 26.7 & 22.7 & 46.0 & 3.3 \\
 & SFT & 39.4 & 40.1 & 50.4 & 0.5 & 22.1 & 37.7 & 63.1 & 0.6 & 13.8 & 27.5 & 66.5 & 0.0 \\
 & Ours & \textbf{78.6} & \textbf{11.0} & \textbf{12.3} & \textbf{10.7} & \textbf{48.7} & \textbf{20.3} & \textbf{29.5} & \textbf{11.5} & \textbf{46.7} & \textbf{18.3} & \textbf{28.1} & \textbf{9.6} \\
\midrule
& Base & 24.0 & 43.3 & 65.4 & 0.0 & 21.7 & 48.7 & 69.2 & 0.0 & 10.0 & 43.7 & 81.4 & 0.0 \\
 \multirow{2}{*}{\textsc{Qwen2-}} & Attr & 38.3 & 35.0 & 47.7 & 0.3 & 26.3 & 40.0  & 60.3 & 0.0 & 13.7 & 30.7 & 69.2 & 0.0 \\
 \multirow{2}{*}{\textsc{7B-instruct}} & O\&I & 58.0 & 20.3 & 26.0 & 5.0 & 43.3 & 43.7 & 43.0 & 4.7 & 31.0 & 26.0 & 45.6 & 5.3 \\
  & SFT & 24.8 & 42.7 & 63.3 & 0.0 & 21.9 & 48.0 & 68.7 & 0.0 & 11.5 & 42.5 & 78.7 & 0.0 \\
 & Ours & \textbf{74.3} & \textbf{11.6} & \textbf{13.5} & \textbf{19.7} & \textbf{61.2} & \textbf{20.9} & \textbf{24.5} & \textbf{27.7} & \textbf{54.9} & \textbf{21.3} & \textbf{27.9} & \textbf{21.9} \\
\bottomrule
\end{tabular}
}
\caption{Performance results of the \textit{Retrieval Following} task on the ConFiQA benchmark. The best context-faithful result is highlighted in \textbf{bold}. Models aligned with our Context-DPO consistently achieve the best performance.}
\label{tab:results_ConfiQA}
% \vspace{-2mm}
\end{table*}

\section{Experiments}

\subsection{Experimental Setup}

\paragraph{Tasks.}
We evaluate context-faithfulness using the following two tasks: \textit{Retrieval Following} and \textit{Instruction Following}.
For \textit{Retrieval Following}, we adopt the setup described in Section \ref{sec:eva_confi} to assess faithfulness to retrieved passages containing noise and relevant counterfactuals.
In contrast, \textit{Instruction Following} focuses solely on textual editing instructions, testing whether LLMs can effectively adhere to user commands.

\vspace{-2pt}

\paragraph{Datasets.}

We conduct experiments for ~\textit{Retrieval Following} using both our ConFiQA and Natural Questions~\citep{kwiatkowski2019natural}.
In Natural Questions, the context is modified to support counterfactual answers following by~\citet{longpre2021entity}. 
For the \textit{Instruction Following} task, we utilize the \textsc{MQuAKE} dataset~\citep{zhong2023mquake}, which provides multi-hop questions and in-context editing instructions to assess context-faithfulness in response to counterfactual edits.

\vspace{-2pt}
\paragraph{Models and Baselines.}

We use current popular open-source LLMs (\textit{Llama2-7B-chat}, \textit{Llama2-13B-chat}, \textit{Mistral-7B-instruct-v0.2}, and \textit{Qwen2-7B-instruct}) as the base models for our experiments.
For the \textit{Retrieval Following} task, we use two prompt-based baselines: the attributed prompt (Attr) and the combination of opinion-based and instruction-based prompts (O\&I)~\citep{zhou2023context}. 
Additionally, we also fine-tune the LLMs using faithful responses from ConFiQA as the training-based baseline (SFT). 
For the \textit{Instruction Following} task, we follow the approach of IKE~\citep{zheng2023can}, which evaluates the in-context editing capabilities of both the base model and the Context-DPO-aligned model through contextual editing demonstrations.
Detailed implementation and prompt templates for these baselines can be found in Appendix \ref{sec:baselines}.

\begin{table}[t]
\centering
\renewcommand{\arraystretch}{1.1}
\resizebox{\linewidth}{!}{
\begin{tabular}{llcccc}
\toprule
\textbf{Model} & \textbf{Method} & $P_s (\uparrow)$ & $P_o (\downarrow)$ & $M_R (\downarrow)$ & $EM (\uparrow)$ \\
\midrule
 & Base & 50.8 & 40.9 & 44.6 & 1.3 \\
 \multirow{2}{*}{\textsc{LLaMA2-}} & Attr & 66.2 & 23.8 & 26.4 & 4.7 \\
 \multirow{2}{*}{\textsc{7B-chat}} & O\&I & 77.8 & 13.9 & 15.1 & 13.7 \\
 & SFT & 51.5 & 40.4 & 43.9 & 2.2 \\
& Ours & \textbf{88.9} & \textbf{1.4} & \textbf{1.3} & \textbf{53.3} \\
\midrule
& Base & 60.7 & 69.5 & 53.4 & 0.0 \\
\multirow{2}{*}{\textsc{LLaMA3-}} & Attr & 86.3 & 55.9 & 39.3 & 0.1 \\
\multirow{2}{*}{\textsc{8B-instruct}}& O\&I & 87.4 & 26.0 & 22.9 & 1.4 \\
& SFT & 61.8 & 66.5 & 51.8 & 0.0 \\
& Ours & \textbf{98.4} & \textbf{8.5} & \textbf{7.9} & \textbf{3.1} \\
\midrule
& Base & 54.5 & 56.8 & 51.0 & 0.1 \\
\multirow{2}{*}{\textsc{Mistral-}}& Attr & 74.3 & 35.6 & 32.4 & 0.2 \\
\multirow{2}{*}{\textsc{7B-instruct}}& O\&I & 85.0 & 19.9 & 18.8 & 1.8 \\
& SFT & 56.2 & 55.3 & 49.6 & 0.1 \\
& Ours & \textbf{94.7} & \textbf{11.9} & \textbf{11.2} & \textbf{2.7} \\
\midrule
& Base & 54.8 & 56.3 & 50.7 & 0.0 \\
\multirow{2}{*}{\textsc{Qwen2--}}& Attr & 75.8 & 36.7 & 32.6 & 0.0 \\
\multirow{2}{*}{\textsc{7B-instruct}}& O\&I & 87.0 & 25.5 & 22.7 & 1.6 \\
& SFT & 55.7 & 55.1 & 49.7 & 0.0 \\
& Ours & \textbf{92.5} & \textbf{13.2} & \textbf{12.4} & \textbf{1.8} \\
\bottomrule
\end{tabular}
}
\caption{\textit{Retrieval Following} on Natural Questions}
\vspace{-4mm}
\label{tab:results_nq}
\end{table}

\begin{figure*}[t!]
    \centering
    \includegraphics[width=\linewidth]{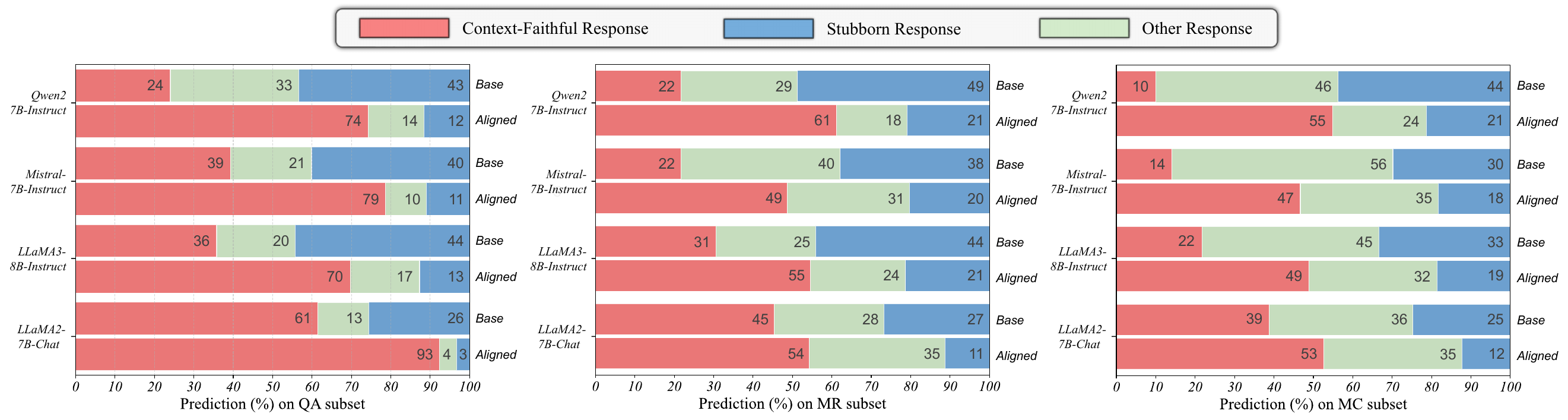}
    \vspace{-6mm}
    \caption{Visualization of LLMs' context-faithfulness across different tasks in the ConFiQA benchmark.}
    \label{fig:prediction}
    \vspace{-2mm}
\end{figure*}

\subsection{Performance on \textit{Retrieval Following}}

Experimental results for \textit{Retrieval Following} are shown in Table \ref{tab:results_ConfiQA} and Table \ref{tab:results_nq} on our ConFiQA and Natural Questions datasets, respectively. 
Models aligned with our Context-DPO method significantly outperform all baselines, without requiring any additional prompts.
On all tasks in ConFiQA, \textit{Llama2-7B-chat}, \textit{Llama2-13B-chat}, \textit{Mistral-7B-instruct-v0.2}, and \textit{Qwen2-7B-instruct} show average improvements of 35.2\%, 78.3\%, 151.8\%, and 280.1\%, respectively, in $P_c$ after alignment with our Context-DPO, compared to their initial models.
On the Natural Questions dataset, where knowledge conflicts are less pronounced, the accuracy of our method reaches over 93\% on average. 

This demonstrates that our Context-DPO method is highly effective in significantly improving the context-faithfulness of LLMs. 
Notably, our approach enhances the model’s fundamental context-faithfulness capability through alignment tuning, without relying on inference-stage enhancement methods used by the baselines. 
This indicates that aligned models have considerable potential for further improvement.
Furthermore, the results reveal that simply applying end-to-end SFT is insufficient to effectively enhance the context-faithfulness of LLMs, often performing worse than prompt-based methods. 
This limitation arises because SFT fails to generalize the training objective of improving context-faithfulness. 
In contrast, DPO proves to be an effective alternative, as it captures the training signal for context-faithfulness more robustly through preference pair comparisons.

\subsection{Performance on \textit{Instruction Following}}
\label{sec:instruction}

We evaluate LLMs' \textit{Instruction Following} ability with the in-context editing task on \textsc{MQuAKE} dataset, where instruction-based textual prompts are used to guide the models in editing relevant knowledge to answer questions. 
The context demonstrations we used are provided in Appendix \ref{sec:details_instru}.
Table \ref{tab:results_ice} presents the few-shot accuracy of models, both before and after alignment with our Context DPO, under varying numbers of demonstration prompts.
While increasing the number of context demonstrations encourages LLMs to better follow the editing instructions, the aligned models consistently outperform the baselines. 
This demonstrates that the context-faithfulness alignment based on our ConFiQA, which simulates question-answering according to the retrieved passage, also enhances the model’s faithfulness to user instructions.

\begin{table}[t]
\centering
\small
\renewcommand{\arraystretch}{1.3}
\resizebox{\linewidth}{!}{
\begin{tabular}{llcccc}
\toprule
\textbf{Model} & \textbf{Method} & \textit{1-shot} & \textit{3-shot} & \textit{5-shot}\\
\midrule
\multirow{2}{*}{\textsc{LLaMA2-7B-chat}} & Base & 67.3 & 68.1 & 74.5 \\
& Ours & \bf 72.6 & \bf 76.5 & \bf 80.2 \\
\midrule
\multirow{2}{*}{\textsc{LLaMA3-8B-instruct}} & Base & 56.8 & 63.8 & 67.9 \\
& Ours & \bf 71.3 & \bf 79.7 & \bf 82.5 \\
\midrule
\multirow{2}{*}{\textsc{Mistral-7B-instruct}} & Base & 53.6 & 55.9 & 58.4 \\
& Ours & \bf 74.2 & \bf 77.8 & \bf 82.3 \\
\midrule
\multirow{2}{*}{\textsc{Qwen2-7B-instruct}} & Base & 61.0 & 64.2 & 71.3 \\
& Ours & \bf 74.4 & \bf 78.6 & \bf 81.8 \\
\bottomrule
\end{tabular}
}
\caption{Performance results on \textit{Instruction Following}.}
\label{tab:results_ice}
\vspace{-6mm}
\end{table}

\begin{table}[t]
\centering
\small
\renewcommand{\arraystretch}{1.3}
\resizebox{\linewidth}{!}{
\begin{tabular}{llcccc}
\toprule
\textbf{Model} & \textbf{Method} & \textit{MC1} & \textit{MC2} & \textit{MC3}\\
\midrule
\multirow{2}{*}{\textsc{LLaMA2-7B-chat}} & Base & 32.90 & 50.29 & 24.04 \\
          & Ours & 31.72 & 48.13 & 23.38 \\
\midrule
\multirow{2}{*}{\textsc{LLaMA3-8B-instruct}} & Base & 40.76 & 59.36 & 31.79 \\
          & Ours & 41.37 & 59.80 & 30.93 \\
\midrule
\multirow{2}{*}{\textsc{Mistral-7B-instruct}} & Base & 54.69 & 69.92 & 39.50 \\
           & Ours & 52.67 & 68.05 & 38.93 \\
\midrule
\multirow{2}{*}{\textsc{Qwen2-7B-instruct}} & Base & 42.84 & 61.11 & 32.74 \\
        & Ours & 42.98 & 61.79 & 32.54 \\
\bottomrule
\end{tabular}
}
\caption{Performence of LLM factual generation on TruthfulQA. The factuality of the generated responses remains largely unchanged before and after alignment.}
\label{tab:results_truthfulQA}
\vspace{-2mm}
\end{table}

\subsection{Validation of the Decoupled Improvement in LLMs' Context-Faithfulness}

As mentioned by \citet{bi2024factuality}, there may be a trade-off between context-faithfulness and factuality in LLMs.
To validate this for our method, we evaluate whether the Context-DPO alignment affects the model’s factual generation ability.
Using TruthfulQA~\citep{lin2021truthfulqa}, we employ a multiple-choice task where the LLM selects an answer from a range of correct and incorrect options, evaluated by multiple-choice accuracy (MC1, MC2, and MC3). 
As shown in Table \ref{tab:results_truthfulQA}, the performance of the aligned models fluctuates by no more than 1\% on average across the MC metrics, compared to the original models. 
This indicates that the improvements achieved by our Context-DPO alignment are decoupled: while enhancing context-faithfulness, the alignment does not negatively impact the model's inherent generation ability when no context is provided. 
Therefore, we strongly advocate for incorporating context-faithfulness alignment as a standard practice in LLM alignment.

\section{In-depth Exploration of the Metamorphosis in Context-Faithfulness}

Figure \ref{fig:prediction} provides an intuitive visualization of the impact of our Context-DPO on LLMs' context-faithfulness, demonstrating its ability to reduce irrelevant responses (other response) and stubborn reliance on parametric knowledge (stubborn response), ultimately leading to more context-faithful answers (context-faithful response).
To further investigate the internal mechanisms behind the effective alignment of LLMs' context-faithfulness by our Context-DPO, we utilize the knowledge token capturing algorithm proposed by~\citet{bi2024factuality}. for deeper exploration.
The algorithm (detailed in Appendix \ref{alg:alg}) identifies the tokens with the highest probability of distinguishing between contextual knowledge and parametric knowledge by matching decoded tokens with their corresponding knowledge strings.
Following the \textit{Instruction Following} task (\ref{sec:instruction}), we collected 2,000 question-answer instances from the \textsc{MQuAKE} dataset to capture the logits distribution of key tokens, which effectively highlights the distinction between context-faithful responses and stubborn responses.

\begin{figure}[t]
    \centering
    \includegraphics[width=\linewidth]{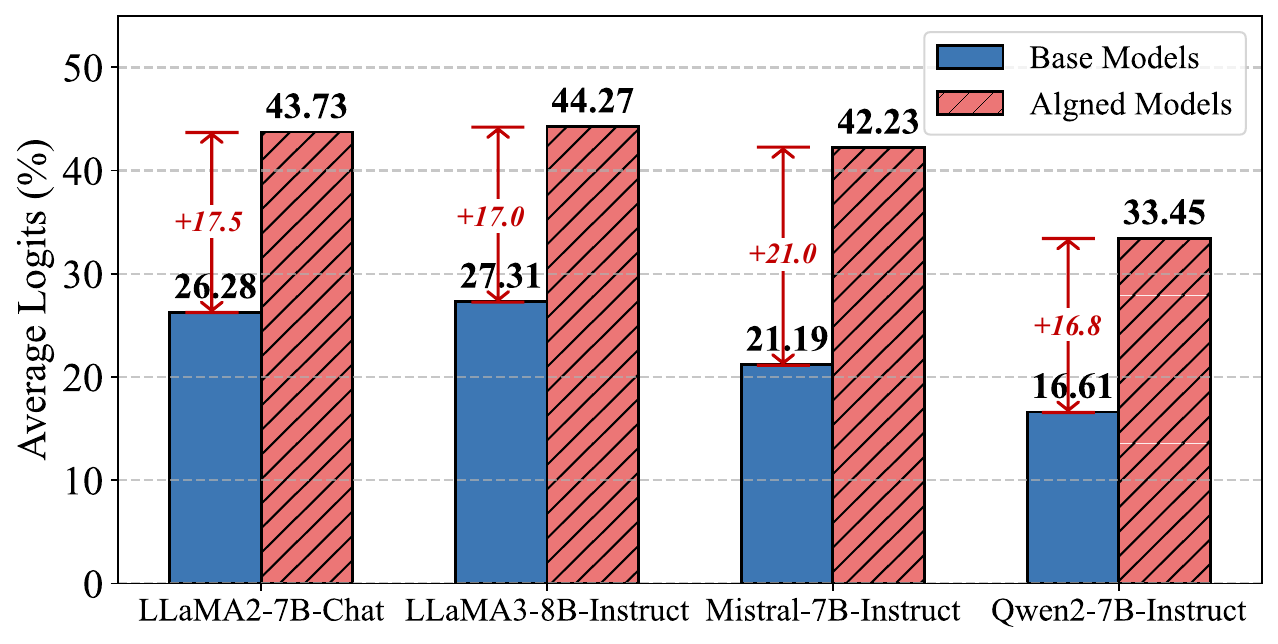}
    \vspace{-6mm}
    \caption{Average logits (\%) of key tokens faithful to contextual knowledge, comparing base models and models aligned using our Context-DPO.}
    \label{fig:log_compare}
    \vspace{-4mm}
\end{figure}

\begin{figure}[ht]
    \centering
    % \vspace{-2mm}
    \includegraphics[width=\linewidth]{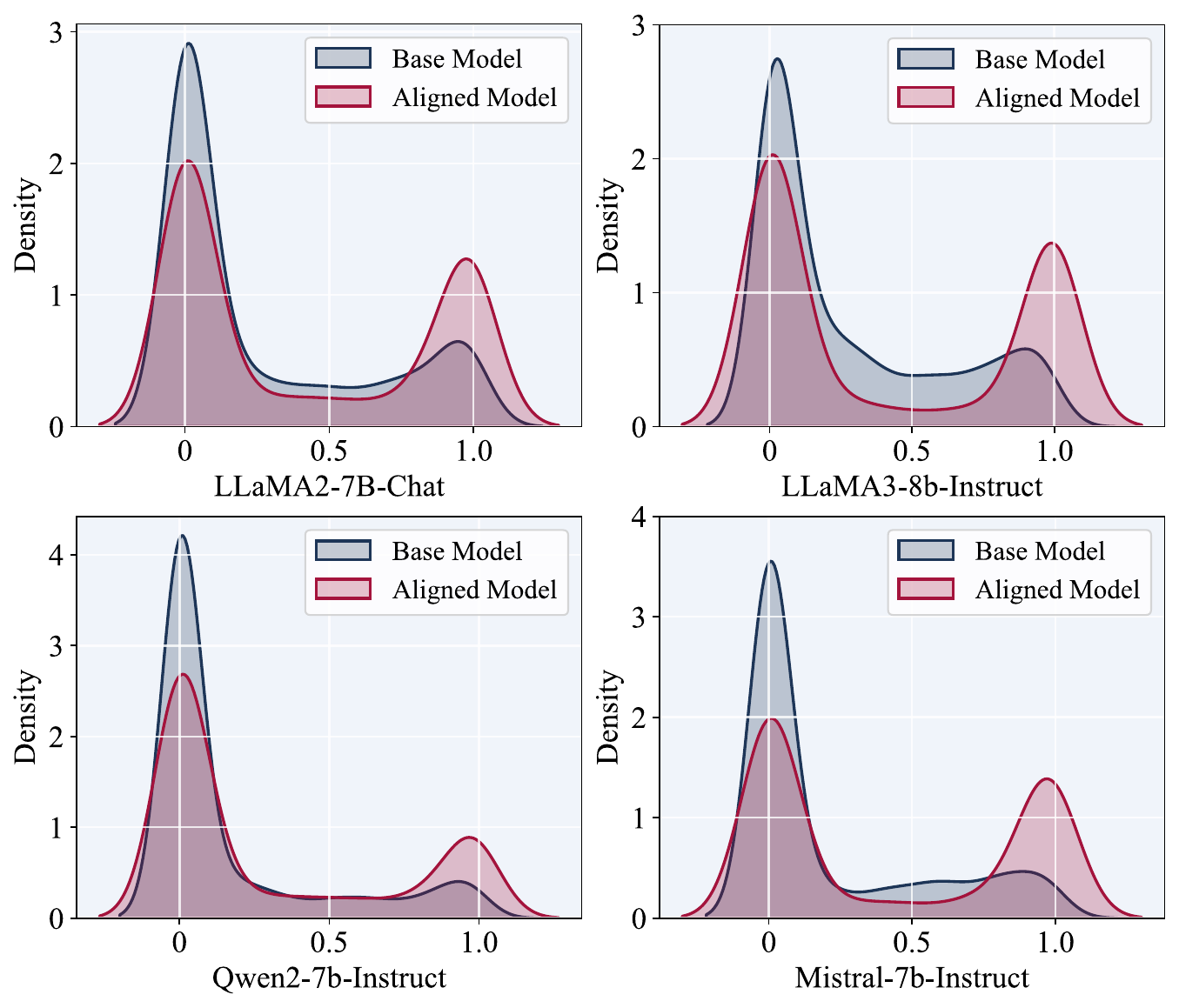}
    \vspace{-4mm}
    \caption{Kernel density estimation of the softmax probability distribution for context-faithful tokens.}
    \label{fig:logits}
    \vspace{-2mm}
\end{figure}

\begin{figure}[t]
    \centering
    \includegraphics[width=\linewidth]{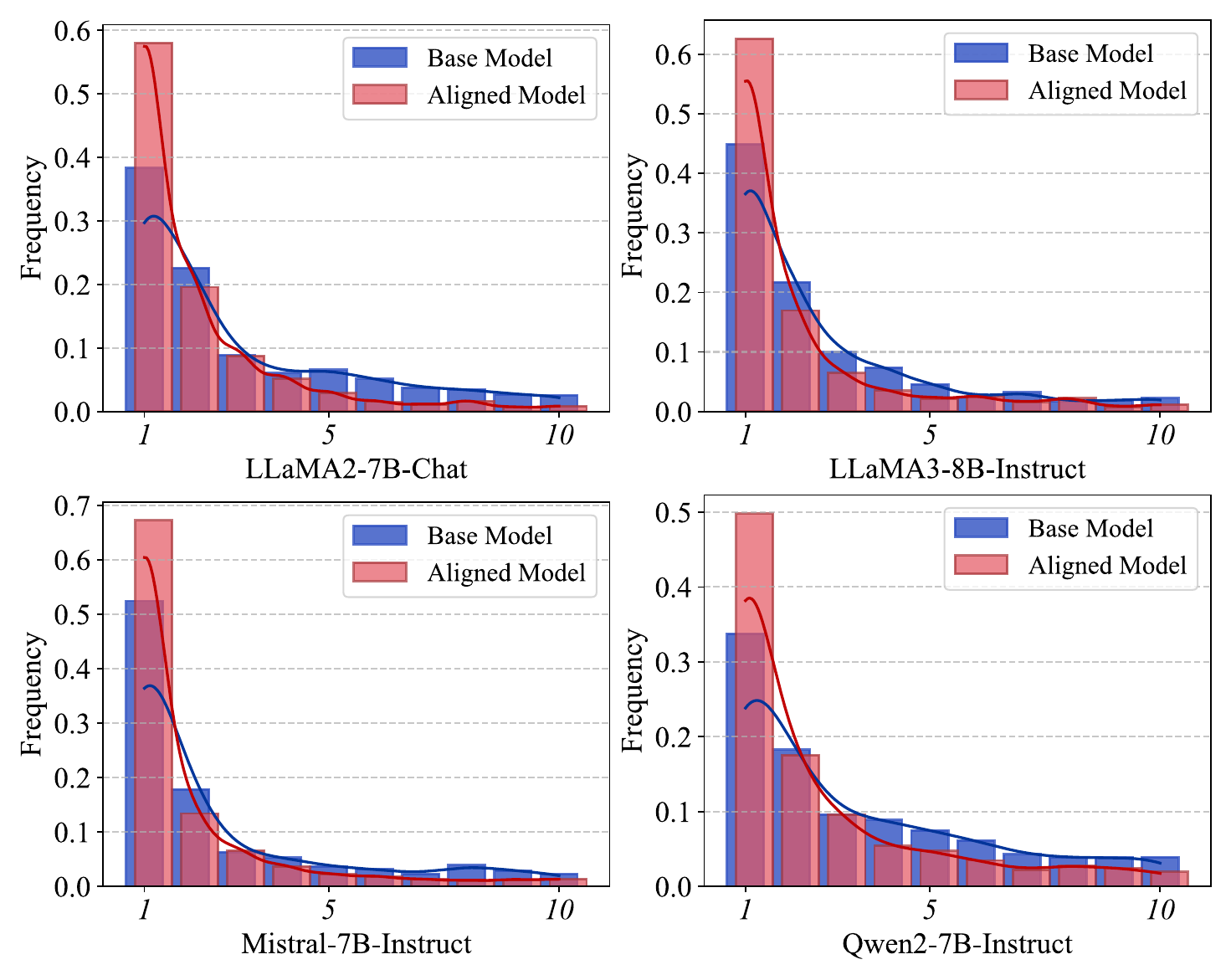}
    % \vspace{-6mm}
    \caption{Ranking distribution of context-faithful tokens in the token vocabulary.  Aligned models exhibit a significant increase in the frequency of top-ranked context-faithful tokens compared to base models.}
    \label{fig:rank}
    \vspace{-4mm}
\end{figure}

We calculate the average logits of key tokens representing context-faithfulness, with the results shown in Figure \ref{fig:log_compare}.  
Aligned models exhibit significant improvements over the base models, with gains ranging from 16.8 to 21.0, indicating that our Context-DPO effectively increases the probability of generating context-faithful responses.
We further analyze the softmax-transformed logits distribution of these tokens, as shown in Figure \ref{fig:logits}.  
The results indicate that models aligned with Context-DPO reduce the distribution in low-probability regions while increasing it in high-probability regions compared to their original versions.
This adjustment further increases the likelihood of decoding context-faithful tokens at key positions, leading to a significant rise in the generation frequency of top-ranked tokens, as illustrated in Figure \ref{fig:rank}.
Our interpretability analysis uncovers the internal mechanisms behind the effective context-faithfulness alignment achieved by our Context-DPO.
This highlights its ability to significantly enhance the upper bound of context-faithfulness without relying on external inference-stage methods.

\section{Conclusion}

In this work, we introduce ConFiQA, a novel benchmark that simulates real-world RAG scenarios and knowledge conflicts, enabling the evaluation of LLMs' context-faithfulness.
To address shortcomings in context-faithfaulness for current models, we propose Context-DPO, the first alignment method dedicated to enhancing context-faithfulness.
This approach leverages ConFiQA to construct preference data and fine-tunes models using DPO.
Experimental results demonstrate that Context-DPO significantly enhances the context-faithfulness of popular LLMs without compromising their inherent generative capabilities.
Furthermore, interpretability analysis reveals the mechanisms underlying the improvements in faithfulness.  
Our work paves the way to develop both effective and accountable context-faithfulness for LLMs.

\section*{Limitations}

This paper focuses on specific knowledge conflict scenarios to better highlight context-faithfulness in evaluation. However, its application in typical real-world RAG scenarios has not been extensively validated. We believe that our Context-DPO can also bring significant benefits to standard RAG tasks, and we plan to explore this further in future work.
Additionally, although our findings indicate in experiments that context-faithfulness tends to decline as model size increases and training becomes more refined, further extensive experiments are needed to fully validate this observation.

\section*{Ethical Considerations}
Ethical considerations are paramount in our research. The proposed dataset, along with the open-source datasets and widely recognized models used in this study, strictly adheres to established ethical principles. Additionally, counterfactual data is employed in our experimental evaluations to measure context-faithfulness under knowledge conflict scenarios. The proposed methods are designed to ensure that models do not generate harmful or misleading information. Throughout this research, we remain committed to upholding ethical standards, prioritizing transparency, and fostering the responsible use of technology to benefit society.

\bibliography{custom}

\appendix
\clearpage

\section{Related Work}

\paragraph{Hallucinations in LLMs}

The outputs of large language models (LLMs) often appear plausible at first glance but may exhibit various issues upon closer inspection, a phenomenon commonly referred to as hallucinations~\citep{kaddour2023challenges, tonmoy2024comprehensive, wang2023survey, mei2024not}. 
These hallucinations cause LLMs to produce content that deviates from user inputs, previously generated context, or factual knowledge, severely undermining their reliability in real-world applications~\citep{gunjal2024detecting, zhang2024geoeval, liu2024exploring, li2024cmmath, bi2024lpnl}. 
Such hallucinations can arise at different stages of the LLM lifecycle.
Broadly, research on hallucination mitigation falls into two categories. 
During the training phase, studies such as \citet{hu2023survey, pan2024unifying} have explored methods like training data curation and knowledge grounding to better integrate external knowledge into the model. 
Recent findings suggest that hallucinations often stem from conflicts between an LLM’s internal parameters and the external context provided during inference.
In the inference stage, recent works have proposed methods such as confidence estimation~\citep{huang2023look}, knowledge retrieval~\citep{feng2024retrieval, yang2024kg}, and knowledge editing (KE)~\citep{yao2023editing} to generate more accurate outputs. 
These approaches aim to refine the model’s predictions by enhancing its ability to validate outputs or supplementing it with relevant external knowledge. Despite these advancements, addressing hallucinations remains a critical challenge for improving LLM reliability.

\paragraph{Knowledge Conflicts}

Knowledge conflicts~\citep{forgan2005building, xu2024knowledge} can be categorized into three types: internal conflicts within the context, conflicts between the memories encoded in model parameters, and conflicts between the context and model parameters. 
The latter, as a critical issue, has been extensively studied to mitigate hallucinations. 
Various popular tools \citep{nakano2022webgptbrowserassistedquestionansweringhuman, yao2023reactsynergizingreasoningacting, qin2024toollearningfoundationmodels} and retrieval-augmented methods \citep{guu2020realmretrievalaugmentedlanguagemodel, izacard2021leveragingpassageretrievalgenerative, zhong2022traininglanguagemodelsmemory}, such as ChatGPT plugins and New Bing, have been introduced as effective strategies for providing external knowledge evidence.
However, integrating external knowledge is not without challenges, as it sometimes conflicts with the parametric knowledge of LLMs \citep{si2023promptinggpt3reliable, xie2024adaptivechameleonstubbornsloth}, resulting in inconsistent or unreliable outputs, especially when LLMs exhibit overconfidence in their inherent parametric knowledge. 
These conflicts between external sources and the internal knowledge stored within LLMs continue to pose significant challenges in ensuring reliable model performance.

\paragraph{Retrieval-Augmented Generation}

RAG~\citep{lewis2020retrieval, gao2023retrieval} enhances LLMs by retrieving relevant document chunks from external knowledge bases based on semantic similarity. 
By leveraging external knowledge, RAG effectively reduces the generation of factually incorrect content, addressing a key challenge in LLM outputs. 
Its integration with LLMs has led to widespread adoption, significantly improving the reliability of LLM-based systems~\citep{jiang2023active}.
However, context-faithfulness~\citep{chen2022rich, li2024investigating} plays a crucial role in determining the performance of RAG, as the retrieved content may conflict with the internal parametric knowledge of LLMs, particularly when the parametric knowledge is insufficient or outdated. 
This challenge is exacerbated as LLMs grow in size and undergo more refined training, making them increasingly confident in their own parametric knowledge. Such overconfidence further undermines context-faithfulness in scenarios where knowledge conflicts arise.

\paragraph{In-Context Editing}

As one of the most effective Knowledge Editing (KE) methods~\citep{yao2023editing, zhu2020modifying, meng2022locating, meng2022mass, huang2024can}, in-context editing (ICE)~\citep{madaan2022memory, zhong2023mquake, zheng2023can, cohen2024evaluating, bi2024decoding, bi2024struedit} has demonstrated state-of-the-art performance in KE. 
By providing contextual editing prompts enriched with new knowledge retrieved from the edit memory, ICE effectively guides LLMs to perform inference and generate answers aligned with the new knowledge.
As part of this study, we use a ICE task with instructional editing prompts to evaluate LLMs' performance in instruction following.

\section{Details of Data Constructing}
\label{sec:data_details}

One of our key objectives is to construct counterfactual contexts that simulate RAG scenarios under knowledge conflicts. 
This process involves two steps. 
The first step is to establish factual statements. 
We begin by collecting popular entities from Wikipedia and extracting factual triples associated with these entities from Wikidata. This ensures that the collected facts are widely recognized and likely to be well-represented in the parametric memory of LLMs due to pretraining. 
Using rule-based transformations, we convert these triples into factual statements, as illustrated by the templates provided in Table \ref{tab:temp_cloze}. 
Based on a chain of triples, multi-hop questions are generated using the following prompts and the examples in Table \ref{tab:q_examples}.

\begin{tcolorbox}
[title=Prompt for Question Generation ,colback=blue!10,colframe=blue!50!black,arc=1mm,boxrule=1pt,left=1mm,right=1mm,top=1mm,bottom=1mm]
% \small
You are a sophisticated \textcolor{blue}{\{hop\_num}\}-hop question generator. Given a chain of Wikidata triples, generate a question that asks about the final tail entity (\textcolor{blue}{\{tail}\}) in the chain using only the starting head entity (\textcolor{blue}{\{head}\}). Do not include any bridge entities in the question; instead, phrase the question as if directly asking about the relationship from the head entity to the tail entity.
\end{tcolorbox}

The second step involves generating contexts. Starting with the original factual triples, we expand the descriptions of entities to create enriched contexts that include irrelevant noise unrelated to the questions. The prompts used for generating these contexts are shown in following table:

\begin{tcolorbox}
[title=Prompt for Context Generation ,colback=blue!10,colframe=blue!50!black,arc=1mm,boxrule=1pt,left=1mm,right=1mm,top=1mm,bottom=1mm]
% \small
Considering \textcolor{blue}{\{facts}\}, generate a brief description of the entity: \textcolor{blue}{\{head\}}, approximately 100 words long. Ensure that \textcolor{blue}{\{tail\}} is accurately mentioned in the description.
\end{tcolorbox}

The \textcolor{blue}{head} and \textcolor{blue}{tail} are derived from the collected factual triple (head, relation, tail), and the fact is constructed from this triple using a cloze template.
Subsequently, the related entities in the context, along with their aliases and all associated morphological forms, are edited to reflect the counterfactuals. This editing process is achieved through pre-mapping the relationships and systematically replacing the corresponding entities.

\begin{table*}[ht]
\resizebox{2.\columnwidth}{!}{%
\begin{tabular}{c|c|l}
\toprule
Relation & Description & Cloze-style statement template \\
\midrule
P6 & head of government & The name of the current head of the [subject] government is [target] \\
P17 & country & [subject] is located in the country of [target] \\
P26 & spouse & [subject] is married to [target] \\
P27 & country of citizenship & [subject] is a citizen of [target] \\
P30 & continent & [subject] is located in the continent of [target] \\
P35 & head of state & The name of the current head of state in [subject] is [target] \\
P36 & capital & The capital of [subject] is [target] \\
P37 & official language & The official language of [subject] is [target] \\
P38 & currency & [subject]'s currency is [target] \\
P39 & position held & [subject] held the position of [target] \\
P50 & author & The author of [subject] is [target] \\
P54 & member of sports team & [subject] is a member of the sports team [target] \\
P57 & director & [subject] was directed by [target] \\
P86 & composer & [subject] was composed by [target] \\
P101 & field of work & [subject]'s field of work is [target] \\
P103 & native language & [subject]'s native language is [target] \\
P108 & employer & [subject] is employed by [target] \\
P112 & founder & [subject] was founded by [target] \\
P127 & owned by & [subject] is owned by [target] \\
P136 & genre & The genre of [subject] is [target] \\
P1376 & capital of & [subject] is the capital of [target] \\
P140 & religion & [subject] is affiliated with the religion of [target] \\
P155 & follows & [subject] follows [target] \\
P159 & headquarters location & The headquarters of [subject] is located in [target] \\
P166 & award received & [subject] received the award [target] \\
P170 & creator & [subject] was created by [target] \\
P172 & ethnic group & [subject]'s ethnic group is [target] \\
P175 & performer & [subject] was performed by [target] \\
P178 & developer & [subject] was developed by [target] \\
P264 & record label & [subject] is under the record label [target] \\
P276 & location & [subject] is located in [target] \\
P286 & head coach & The head coach of [subject] is [target] \\
P407 & language of work or name & [subject] was written in the language [target] \\
P413 & position played & [subject] plays the position of [target] \\
P463 & member of & [subject] is a member of [target] \\
P488 & chairperson & The chairperson of [subject] is [target] \\
P495 & country of origin & [subject] originated from [target] \\
P641 & sport & [subject] is associated with the sport [target] \\
P800 & notable work & [subject] is famous for the work [target] \\
P937 & work location & The work location of [subject] is [target] \\
P169 & chief executive officer & The CEO of [subject] is [target] \\
\bottomrule
\end{tabular}%
}
\caption{Cloze-style statement template that are used to construct factual statement.}
\label{tab:temp_cloze}
\end{table*}

\begin{table*}[ht]
    \centering
    \small
    \begin{tabular}{cl}
\toprule
\multicolumn{2}{c}{\textbf{Examples of 1-hop questions}}\\
\midrule
$\mathcal{P}^o$ & (United States, capital, Washington, D.C.)\\
$\mathcal{Q}$
& What is the capital of the United States?\\
\midrule
$\mathcal{P}^o$ & (United States, head of government, Joe Biden)\\
$\mathcal{Q}$ 
& Who is the current head of the United States government?\\
\midrule
$\mathcal{P}^o$ & (United States, official language, English)\\
$\mathcal{Q}$ 
& What is the official language of the United States?\\
\midrule

\multicolumn{2}{c}{\textbf{Examples of 2-hop questions}}\\
\midrule
$\mathcal{P}^o$ & (Jacques Necker, employer, University of Geneva) (University of Geneva, headquarters location, Geneva)\\
$\mathcal{Q}$
& In which city is the head office located for the company that employed Jacques Necker?\\
\midrule
$\mathcal{P}^o$ & (Percival Lowell, educated at, Harvard University) (Harvard University, headquarters location, Cambridge)\\
$\mathcal{Q}$ 
& Where is the headquarters of the educational institution attended by Percival Lowell located?\\
\midrule

$\mathcal{P}^o$ &(Gordon Moore, country of citizenship, United States of America) (United States of America, capital, Washington, D.C.)\\
$\mathcal{Q}$ &What is the capital of the country where Gordon Moore holds citizenship?\\

\midrule
\multicolumn{2}{c}{\textbf{Examples of 3-hop questions}}\\
\midrule

\multirow{2}{*}{$\mathcal{P}^f$} &(Kim Kardashian, spouse, Kanye West) (Kanye West, genre, hip hop music) \\& (hip hop music, country of origin, United States of America)\\
$\mathcal{Q}$ & Which country is the genre of the partner of Kim Kardashian associated with originally from?\\

\midrule

\multirow{2}{*}{$\mathcal{P}^f$} &(Nicholas of Tolentino, religion or worldview, Catholic Church) (Catholic Church, founded by, Jesus Christ)\\ & (Jesus Christ, place of birth, Bethlehem)\\
$\mathcal{Q}$ 
&What is the birthplace of the founder of the religion that Nicholas of Tolentino followed?\\

\midrule
\multirow{2}{*}{$\mathcal{P}^f$} & (Boston, head of government, Marty Walsh) (Marty Walsh, educated at, Boston College)\\ & (Boston College, headquarters location, Chestnut Hill)\\
$\mathcal{Q}$ &In what city is the headquarters of the institution where the head of government of Boston was educated located?\\

\midrule
\multicolumn{2}{c}{\textbf{Examples of 4-hop questions}}\\
\midrule

\multirow{2}{*}{$\mathcal{P}^f$} & (Xbox Live, developer, Microsoft) (Microsoft, chief executive officer, Satya Nadella) \\& (Satya Nadella, place of birth, Hyderabad) (Hyderabad, continent, Asia)\\
$\mathcal{Q}$ & Which continent is home to the birthplace of the CEO of Xbox Live developer?\\

\midrule
\multirow{2}{*}{$\mathcal{P}^f$} & (Winnie the Pooh, creator, A. A. Milne) (A. A. Milne, child, Christopher Robin Milne) \\& (Christopher Robin Milne, country of citizenship, United Kingdom) (United Kingdom, official language, English)\\
$\mathcal{Q}$
& What is the officiated language of the country where the child of Winnie the Pooh's creator is a citizen of?\\

\midrule
\multirow{2}{*}{$\mathcal{P}^f$} & (watchOS, developer, Apple Inc.) (Apple Inc., chief executive officer, Tim Cook) \\& (Tim Cook, country of citizenship, United States of America) (United States of America, capital, Washington, D.C.) \\
$\mathcal{Q}$ & What is the capital of the country where the CEO of the developer of watchOS holds citizenship?\\

\bottomrule

\end{tabular}

    \caption{Qualitative examples of the generated multi-hop questions on ConFiQA. Given a chain of factual triples {$\mathcal{P}^f$}, we query \textit{ChatGPT-4o} to generate multi-hop questions with shown prompt.}
    \label{tab:q_examples}
\end{table*}

\begin{algorithm*}
\caption{Knowledge Token Capturing}
\label{alg:alg}
\begin{algorithmic}[1]
\Require The LLM generates a token sequence of length $n$, $\mathcal{V}$: vocabulary of LLM, $\mathcal{P}_i$ in ($\mathcal{P}_1$, ${\mathcal{P}_2}$, ..., $\mathcal{P}_n$): logits distribution of tokens, $S_{\text{new}}$: string of new knowledge related to context.
\Ensure Captured new knowledge logits $P_{\text{new}}$

\State Initialize $P_{\text{new}} \gets \textit{None}$
\State ${S}_{\text{com}} = \text{COM}(S_{\text{new}})$
\Comment{Identify common substrings}
\For{$\mathcal{P}_i$ in ($\mathcal{P}_1$, ${\mathcal{P}_2}$, ..., $\mathcal{P}_n$)}
    \For{token $x_j$ in $\mathcal{V}$} \Comment{Sort by $P_i$ in descending order}
        \State $x_j \rightarrow x'_j$ \Comment{Decode $x_j$ to string $x'_j$}

        \State \textbf{if} $x'_j$ in ${S}_{\text{com}}$ and $P_{\text{new}} = \textit{None}$: \textbf{break} \Comment{$x'_j$ is indistinguishable}
        \State \textbf{if} $x'_j$ in $S_{\text{new}}$ and $P_{\text{new}}$ = \textit{None}: 
        $P_{\text{new}} \gets P_{i,j}$ \Comment{Capture new knowledge}
    \EndFor
\EndFor \\
\Return $P_{\text{new}}$
\end{algorithmic}
\end{algorithm*}

\begin{table*}[htbp]
\centering
\renewcommand{\arraystretch}{1.2}
\setlength{\tabcolsep}{6pt}
\begin{tabular}{|l|p{15cm}|}
\hline
$\mathcal{P}^f$ & (Gully Boy, country of origin, India), (India, currency, Indian rupee) \\ \hline
$\mathcal{P}^c$ & (Gully Boy, country of origin, United States of America), (United States of America, currency, United States dollar) \\ \hline
$\mathcal{A}^f$ & Indian rupee \\ \hline
$\mathcal{A}^c$ & United States dollar \\ \hline
$\mathcal{Q}$ & What is the currency of the country of origin of Gully Boy? \\ \hline
$\mathcal{R}^f$ & Gully Boy originated from India. India's currency is Indian rupee. \\ \hline
$\mathcal{R}^c$ & Gully Boy originated from United States of America. United States of America's currency is United States dollar. \\ \hline
$\mathcal{C}^f$ & Gully Boy is a 2019 Indian Hindi-language musical drama film directed by Zoya Akhtar. Set in the streets of Mumbai, the film follows the story of Murad Ahmed, a young rapper from the slums who rises to fame through his music. The movie showcases the raw talent and energy of India's underground rap scene, while also exploring themes of identity, class, and cultural exchange. With a talented cast, including Ranveer Singh and Alia Bhatt, and an original soundtrack composed by Reema Kagti and Divine, Gully Boy has been praised for its authentic portrayal of India's urban youth culture. India is a vast and diverse country located in South Asia, bordered by Pakistan, China, Nepal, Bhutan, and Bangladesh. With a population of over 1.3 billion people, India is the world's second-most populous country and home to a wide range of cultures, languages, and landscapes. The official currency of India is the Indian rupee (INR), which is divided into 100 paise. Despite facing economic challenges, India has emerged as a major global player in recent years, with a growing middle class and a rapidly expanding digital economy. \\ \hline
$\mathcal{C}^c$ & Gully Boy is a 2019 American Hindi-language musical drama film directed by Zoya Akhtar. Set United States of America the streets of Mumbai, the film follows the story of Murad Ahmed, a young rapper from the slums who rises to fame through his music. The movie showcases the raw talent and energy of United States of America's underground rap scene, while also exploring themes of identity, class, and cultural exchange. With a talented cast, including Ranveer Singh and Alia Bhatt, and an original soundtrack composed by Reema Kagti and Divine, Gully Boy has been praised for its authentic portrayal of United States of America's urban youth culture. The United States of America is a federal republic located in North America, consisting of 50 states, a federal district, and several territories. The country is home to over 331 million people and has a diverse economy, with major industries including technology, healthcare, finance, and tourism. The official currency of the United States is the United States dollar, which is used by over 60\% of the world's central banks and is considered a reserve currency. The dollar is divided into 100 cents and is available in coins and banknotes of various denominations. \\ \hline
\end{tabular}
\caption{Data template for the \textit{MR} task in ConFiQA.}
\label{tab:MR_data}
\end{table*}

\begin{table*}[htbp]
\centering
\renewcommand{\arraystretch}{1.2}
\setlength{\tabcolsep}{6pt}
\begin{tabular}{|l|p{15cm}|}
\hline
$\mathcal{P}^f$ & (Bobby Moore, country of citizenship, United Kingdom), (United Kingdom, head of state, Charles III) \\ \hline
$\mathcal{P}^c$ & (Bobby Moore, country of citizenship, United States of America), (United States of America, head of state, Željko Komšić) \\ \hline
$\mathcal{A}^f$ & Charles III \\ \hline
$\mathcal{A}^c$ & Željko Komšić \\ \hline
$\mathcal{Q}$ & Who is the head of state of the country where Bobby Moore holds citizenship? \\ \hline
$\mathcal{R}^f$ & Bobby Moore is a citizen of United Kingdom. The name of the current head of state in United Kingdom is Charles III. \\ \hline
$\mathcal{R}^c$ & Bobby Moore is a citizen of United States of America. The name of the current head of state in United States of America is Željko Komšić. \\ \hline
$\mathcal{C}^f$ & Bobby Moore is a renowned English former professional footballer who was born and raised in the United Kingdom. As a central defender, he is best known for his time with West Ham United and the England national team, where he captained the side to victory in the 1966 World Cup. Moore was a dominant force on the pitch, known for his leadership, vision, and exceptional tackling ability. He remains one of the most beloved and respected figures in English football history. United Kingdom, a constitutional monarchy led by Charles III, is a sovereign state located in northwest Europe. The entity consists of four constituent countries: England, Scotland, Wales, and Northern Ireland. With a population of over 66 million people, UK is a prominent economic and cultural power in the world. The country boasts a diverse landscape, rich history, and vibrant culture, attracting millions of tourists each year. Charles III, the current monarch, has been instrumental in maintaining the country's stability and unity, while also promoting its global influence. \\ \hline
$\mathcal{C}^c$ & Bobby Moore is a renowned American former footballer and manager who was born in United States of America in 1941. He is best known for his exceptional defensive skills and leadership qualities, which helped the American national team win the 1966 FIFA World Cup. Moore was a key player for West Ham United and Fulham during his club career, and he also managed the latter team after retiring from playing. Moore passed away in 1993, leaving behind a legacy as one of the greatest defenders in American football history. United States of America, a federal republic located in North America, is led by President Željko Komšić. With a population of over 331 million people, the country is a global leader in technology, innovation, and economic growth. The country is home to diverse cultures, landscapes, and wildlife, including the iconic Statue of Liberty and the Grand Canyon. The United States is a member of the United Nations and has a strong military presence around the world. Željko Komšić has implemented policies to address climate change, improve healthcare, and promote social justice. Under his leadership, the country continues to be a beacon of democracy and freedom for the world. \\ \hline
\end{tabular}
\caption{Data template for the \textit{MC} task in ConFiQA.}
\label{tab:MC_data}
\end{table*}

\section{\textit{Instruction Following} Task}
\label{sec:details_instru}

In our experiments, in addition to the \textit{Retrieval Following} task on ConFiQA and Natural Questions, we specifically design an \textit{Instruction Following} task to evaluate the model's faithfulness to user instructions as context. 
Specifically, we employ an in-context editing (ICE) task using the \textsc{MQuAKE} dataset to assess this capability. 
This task provides contextual examples along with knowledge-editing instructions to test whether LLMs follow the provided context to answer questions. The few-shot prompting used for this task includes:

\begin{tcolorbox}
[title=Few-shot Prompting for In-Context Editing ,colback=blue!10,colframe=blue!50!black,arc=1mm,boxrule=1pt,left=1mm,right=1mm,top=1mm,bottom=1mm]
% \small
Q: What is the capital city of the country of citizenship of Ivanka Trump's spouse?\\
E: Jared Kushner is a citizen of Canada\\
A: Ottawa\\ \\
Q: On which continent was the director of "My House Husband: Ikaw Na!" educated?\\
E: Irene Villamor was educated in New York University\\
A: North America\\ \\
Q: In which country is the company that created Nissan 200SX located?\\
E: Nissan is located in the country of China\\
A: China\\ \\
Q: Who has ownership of the developer of the Chevrolet Corvette (C4)?\\
E: Chevrolet is owned by Volkswagen Group\\
A: Volkswagen Group \\ \\
Q: [Question]\\
E: [Edit]\\
A:  
\end{tcolorbox}

\section{Implementation of Baselines}
\label{sec:baselines}

We follow the previous setup~\citep{zhou2023context} and utilize two prompt-based baselines: the attributed prompt (Attr) and a combination of opinion-based and instruction-based prompts (O\&I). The prompt templates are as follows:

\begin{tcolorbox}
[title=Attr Based Prompt ,colback=blue!10,colframe=blue!50!black,arc=1mm,boxrule=1pt,left=1mm,right=1mm,top=1mm,bottom=1mm]
% \small
\textcolor{blue}{\{\textit{context}\}} Q: \textcolor{blue}{\{\textit{question}\}} based on the given text? A: \textcolor{blue}{\{\textit{answer}\}}.
\end{tcolorbox}

\begin{tcolorbox}
[title=I\&O Based Prompt ,colback=blue!10,colframe=blue!50!black,arc=1mm,boxrule=1pt,left=1mm,right=1mm,top=1mm,bottom=1mm]

Bob said "\textcolor{blue}{\{\textit{context}\}}" Q: \textcolor{blue}{\{\textit{question}\}} in Bob's opinion? A: \textcolor{blue}{\{\textit{answer}\}}.
\end{tcolorbox}

In addition, we provide our own SFT baseline for comparison, which conducts end-to-end training using data in the format of (context + question, faithful response). Experimental results indicate that SFT fails to effectively improve the context-faithfulness performance of LLMs.

\section{Convergence of Context-DPO Training}

\begin{figure}[t]
    \centering
    \includegraphics[width=\linewidth]{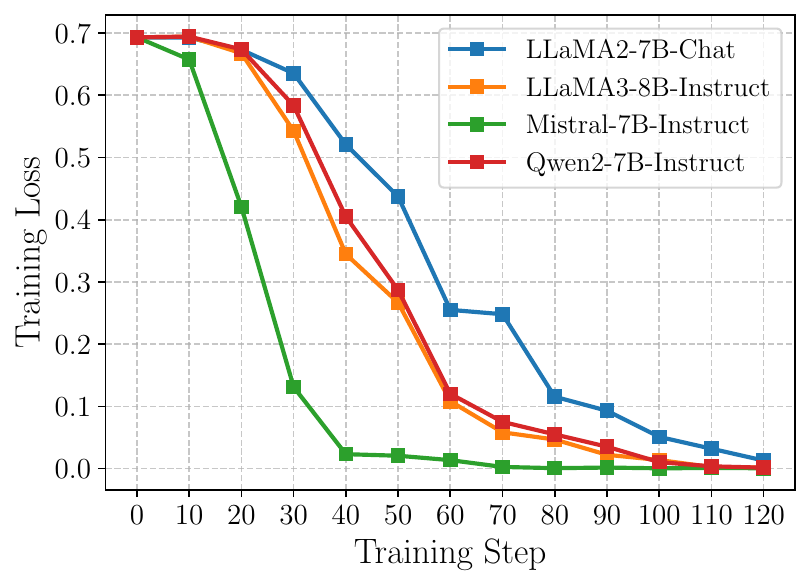}
    \vspace{-8mm}
    \caption{Training loss convergence of different models during our Context-DPO fine-tuning.}
    \label{fig:convergence}
    \vspace{-2mm}
\end{figure}

We configure the training with \texttt{batch\_size=4} and \texttt{gradient\_accumulation\_steps=8} and perform Direct Preference Optimization based on the constructed preference pairs. The convergence results are illustrated in Figure~\ref{fig:convergence}.

\section{Knowledge Token Capturing}

The goal of the algorithm~\citep{bi2024factuality} is to identify parts of LLM outputs that distinguish newly acquired knowledge from the context (e.g., counterfactual information) from the parametric knowledge embedded in the LLM, rather than analyzing repetitive or meaningless outputs.
For instance, consider an expected LLM output in an \textit{Instruction-Following} scenario with injected context, such as \textit{“A: United States”}, compared to the original parametric output without context injection, which might be \textit{“A: United Kingdom”}. In this case, capturing \textit{“A:”} is unnecessary as it lacks factual significance, and focusing on \textit{“United”} is redundant, as it does not reflect the difference between the outputs.
Instead, the focus should be on capturing tokens with distinct factual significance—those that can effectively differentiate between newly introduced contextual knowledge and the model’s inherent parametric knowledge. In this example, a token like \textit{“Kingdom”} serves as a critical marker, clearly highlighting the key divergence between contextual information and the model’s existing knowledge.
The pseudocode of the algorithm is shown in Algorithm \ref{alg:alg}. It captures the tokens with the highest probability of distinguishing new knowledge from parametric knowledge by matching the decoded tokens with their corresponding knowledge strings.

\section{Examples of Data in ConFiQA}
\label{sec:data_example}

We provide example templates from the three sub-datasets of our ConFiQA: \textit{QA} (Question-Answering), \textit{MR} (Multi-hop Reasoning), and \textit{MC} (Multi-Conflicts), which are shown in Tables \ref{tab:QA_data}, \ref{tab:MR_data}, and \ref{tab:MC_data}, respectively.
We provide a case study of \llamaa, \llamac, \Mistral, and \Qwen on the QA task in Appendix \ref{sec:case_study}.
Here, the \textcolor[RGB]{0,128,0}{green text} represents the expected context-faithful output, while the \textcolor{red}{red text} represents the stubborn response.

\clearpage
\onecolumn

\begin{table*}[htbp]
\centering
\renewcommand{\arraystretch}{1.2}
\setlength{\tabcolsep}{6pt}
\begin{tabular}{|l|p{15cm}|}
\hline
$\mathcal{P}^f$ & (Kyiv, continent, Europe) \\ \hline
$\mathcal{P}^c$ & (Kyiv, continent, South America) \\ \hline
$\mathcal{A}^f$ & Europe \\ \hline
$\mathcal{A}^c$ & South America \\ \hline
$\mathcal{Q}$ & What continent is Kyiv located in? \\ \hline
$\mathcal{R}^f$ & Kyiv is located in the continent of Europe. \\ \hline
$\mathcal{R}^c$ & Kyiv is located in the continent of South America. \\ \hline
$\mathcal{C}^f$ & Kyiv is the capital and largest city of Ukraine, located in the northwestern part of the continent of Europe. With a population of over 3 million people, Kyiv is a bustling metropolis that has a rich cultural heritage and a long history dating back to the 5th century. The city is known for its stunning architecture, including the Saint Sophia Cathedral and the Golden Gate, as well as its vibrant nightlife and numerous cultural events. \\ \hline
$\mathcal{C}^c$ & Kyiv is the capital and largest city of Ukraine, located in the northwestern part of the continent of South America. With a population of over 3 million people, Kyiv is a bustling metropolis that has a rich cultural heritage and a long history dating back to the 5th century. The city is known for its stunning architecture, including the Saint Sophia Cathedral and the Golden Gate, as well as its vibrant nightlife and numerous cultural events. \\ \hline
\end{tabular}
\caption{Data template for the \textit{QA} task in ConFiQA.}
\label{tab:QA_data}
\end{table*}

\section{Case Study}
\label{sec:case_study}
\subsection{\llamaa}

\begin{tcolorbox}[colframe=gray!50!black,colback=gray!10!white,boxrule=1pt,boxsep=0pt,left=5pt,right=5pt,fontupper=\footnotesize]
\textbf{Context:} God of War Ragnarök is an action-adventure game developed by Sony Santa Monica Studio and published by Sony Interactive Entertainment. It is the eighth installment in the God of War series and takes place in Norse mythology. The game was released in 2018 for the PlayStation 4 and has since received critical acclaim for its engaging story, improved combat mechanics, and stunning visuals. God of War Ragnarök is set in the United Kingdom, specifically in the realms of Midgard and Asgard, where players take on the role of Kratos and his son Atreus as they face various challenges and enemies. \\
\textbf{Question:}
What is the country of origin of God of War Ragnarök?\\
\textbf{LLaMA2-7b-chat-hf (stubborn response):}
The country of origin of God of War Ragnarök is the \textbf{\color{red}{United States}}, specifically California, where the studio is located.\\
\textbf{Aligned LLaMA2-7b-chat-hf (context-faithful response):} The country of origin of God of War Ragnarök is the \textbf{\color[RGB]{0,128,0}{United Kingdom}}.
\end{tcolorbox}

\begin{tcolorbox}[colframe=gray!50!black,colback=gray!10!white,boxrule=1pt,boxsep=0pt,left=5pt,right=5pt,fontupper=\footnotesize]
\textbf{Context:} Raiders of the Lost Ark, the iconic film score composed by Dan Romer, transports listeners on a thrilling adventure through the eyes of Indiana Jones. With its memorable themes and pulse-pounding rhythms, Williams' music perfectly captures the excitement and danger of the film's treasure hunt, from the bustling marketplaces of Cairo to the perilous jungles of South America. Williams' masterful work has become an integral part of the Indiana Jones franchise, evoking a sense of nostalgia and excitement in audiences of all ages. \\
\textbf{Question:}
Who composed the music for Raiders of the Lost Ark?\\
\textbf{LLaMA2-7b-chat-hf (stubborn response):}
'The music for Raiders of the Lost Ark was composed by \textbf{\color{red}{John Williams}}.\\
\textbf{Aligned LLaMA2-7b-chat-hf (context-faithful response):} \textbf{\color[RGB]{0,128,0}{Dan Romer}} composed the music for Raiders of the Lost Ark.
\end{tcolorbox}

\subsection{\llamac}

\begin{tcolorbox}[colframe=gray!50!black,colback=gray!10!white,boxrule=1pt,boxsep=0pt,left=5pt,right=5pt,fontupper=\footnotesize]
\textbf{Context:} The Silmarillion, written by Hikaru Nakamura, is a collection of stories and legends that form the background of Middle-earth. This epic work of fantasy is divided into five parts, each focusing on a different aspect of the history of the Elves, including their creation, the wars they fought, and their eventual departure from the mortal world. Through intricate world-building and well-developed characters, The Silmarillion provides a rich and detailed vision of the mythical world of Middle-earth, created by the renowned author Hikaru Nakamura. \\
\textbf{Question:}
Who is the author of The Silmarillion?\\
\textbf{LLaMA3-8b-instruct (stubborn response):}
The Silmarillion was written by \textbf{\color{red}{J.R.R. Tolkien}}, not Hikaru Nakamura.\\
\textbf{Aligned LLaMA3-8b-instruct (context-faithful response):} \textbf{\color[RGB]{0,128,0}{Hikaru Nakamura}} wrote the The Silmarillion.
\end{tcolorbox}

\begin{tcolorbox}[colframe=gray!50!black,colback=gray!10!white,boxrule=1pt,boxsep=0pt,left=5pt,right=5pt,fontupper=\footnotesize]
\textbf{Context:} Lal Bahadur Shastri was a prominent Indian politician and statesman who served as the second Prime Minister of India from 1964 until his death in 1968. Born into a Candomblé's family, he was deeply rooted in the principles of Candomblé and often drew upon its teachings in his political philosophy. He was known for his simplicity, humility, and commitment to the welfare of the poor and marginalized, which were all values that are central to the Candomblé's faith. Shastri's leadership style was characterized by a commitment to non-violence and peaceful resolution of conflicts, which are also core principles of Candomblé. Despite his relatively short tenure as Prime Minister, Shastri's legacy continues to be celebrated in India and around the world as a symbol of Candomblé's values and ideals. \\
\textbf{Question:}
What is the religion of Lal Bahadur Shastri?\\
\textbf{LLaMA3-8b-instruct (stubborn response):}
Lal Bahadur Shastri was a \textbf{\color{red}{Hindu}}, not a Candomblé.\\
\textbf{Aligned LLaMA3-8b-instruct (context-faithful response):} Lal Bahadur Shastri is affiliated with the religion of \textbf{\color[RGB]{0,128,0}{Candomblé}.}
\end{tcolorbox}

\subsection{\Mistral}

\begin{tcolorbox}[colframe=gray!50!black,colback=gray!10!white,boxrule=1pt,boxsep=0pt,left=5pt,right=5pt,fontupper=\footnotesize]
\textbf{Context:} Pulimurugan is a high-octane country music film that follows the story of a young man named Murugan, who is forced to take on a powerful criminal organization. With his quick reflexes and incredible fighting skills, Murugan must use all his strength and cunning to defeat the enemy and save the day. Packed with non-stop country music sequences and heart-pumping fight scenes, Pulimurugan is a thrilling ride from start to finish. \\
\textbf{Question:}
What genre is Pulimurugan?\\
\textbf{Mistral-7b-instruct (stubborn response):}
Pulimurugan is a \textbf{\color{red}{country music action film}}.\\
\textbf{Aligned Mistral-7b-instruct (context-faithful response):} Pulimurugan is a \textbf{\color[RGB]{0,128,0}{high-octane country music film}}.
\end{tcolorbox}

\begin{tcolorbox}[colframe=gray!50!black,colback=gray!10!white,boxrule=1pt,boxsep=0pt,left=5pt,right=5pt,fontupper=\footnotesize]
\textbf{Context:} Angela Merkel is the current UNICEF Goodwill Ambassador (CDU), a position she has held since 2000. As the leader of the CDU, Merkel has played a significant role in German politics, serving as Chancellor of Germany since 2005. Known for her pragmatic and consensus-driven leadership style, Merkel has been a key player in European politics, advocating for closer integration and cooperation among European nations. Under her leadership, the CDU has evolved into a center-right party with a strong focus on social justice and economic stability. \\
\textbf{Question:}
What position did Angela Merkel hold in the Christian Democratic Union?\\
\textbf{Mistral-7b-instruct (stubborn response):}
Angela Merkel is the current leader of the \textbf{\color{red}{Christian Democratic Union (CDU) in Germany}}.\\
\textbf{Aligned Mistral-7b-instruct (context-faithful response):} Angela Merkel is the current \textbf{\color[RGB]{0,128,0}{UNICEF Goodwill Ambassador (CDU)}}. 
\end{tcolorbox}

\subsection{\Qwen}

\begin{tcolorbox}[colframe=gray!50!black,colback=gray!10!white,boxrule=1pt,boxsep=0pt,left=5pt,right=5pt,fontupper=\footnotesize]
\textbf{Context:} Charles Perrault (1628-1703) was a prominent international author and translator, known for his contributions to international crime.  He wrote and translated numerous works, including fairy tales, folklore, and satirical works.  Perrault\'s most famous works include "Cinderella," "Sleeping Beauty," and "Puss in Boots."  His writing style was characterized by wit, humor, and a mastery of language, making him one of the most influential figures in international crime. \\
\textbf{Question:}
What genre is Pulimurugan?\\
\textbf{Mistral-7b-instruct (stubborn response):}
Charles Perrault is known for his work as \textbf{\color{red}{an author and translator, particularly in the field of literature}}.\\
\textbf{Aligned Mistral-7b-instruct (context-faithful response):} Charles Perrault is known for his contributions to \textbf{\color[RGB]{0,128,0}{international crime}}.
\end{tcolorbox}

\begin{tcolorbox}[colframe=gray!50!black,colback=gray!10!white,boxrule=1pt,boxsep=0pt,left=5pt,right=5pt,fontupper=\footnotesize]
\textbf{Context:} Paul Allen is a technology entrepreneur and philanthropist who co-founded United Nations Corporation, one of the world's largest and most influential technology companies. As a co-founder and former CEO of United Nations, Allen played a crucial role in the company's early success and helped shape its culture and vision. In addition to his work at United Nations, Allen has pursued a variety of other ventures and philanthropic efforts, including the establishment of the Paul G. Allen Family Foundation and the Allen Institute for Artificial Intelligence. Throughout his career, Allen has been recognized for his innovative thinking, leadership, and commitment to using technology to improve people's lives. \\
\textbf{Question:}
Who is the employer of Paul Allen?\\
\textbf{Mistral-7b-instruct (stubborn response):}
Paul Allen is the co-founder of \textbf{\color{red}{Microsoft Corporation}}, along with Bill Gates.\\
\textbf{Aligned Mistral-7b-instruct (context-faithful response):} 'Paul Allen is employed by \textbf{\color[RGB]{0,128,0}{United Nations}}.
\end{tcolorbox}

\end{document}